\documentclass[sigconf,nonacm]{acmart}
\AtBeginDocument{%
  \providecommand\BibTeX{{%
    \normalfont B\kern-0.5em{\scshape i\kern-0.25em b}\kern-0.8em\TeX}}}

\copyrightyear{2023}
\acmYear{2023}
\setcopyright{none}
\acmConference{Proceedings of the 1st workshop on Social Robots Personalisation: At the crossroads between engineering and humanities}{March 13, 2023}{Stockholm, SE}
\acmBooktitle{Proceedings of the 1st workshop on Social Robots Personalisation: At the crossroads between engineering and humanities, March 13, 2023, Stockholm, Sweden}

\begin{document}

\title{Factors that Affect Personalization of Robots for Older Adults}

\author{Laura Stegner}
\orcid{0000-0003-4496-0727}
\affiliation{%
  \institution{University of Wisconsin--Madison}
  \city{Madison}
  \state{Wisconsin}
  \country{United States}
  \postcode{53706}
}
\email{stegner@wisc.edu}

\author{Emmanuel Senft}
\orcid{0000-0001-7160-4352}
\affiliation{%
  \institution{Idiap Research Institute}
  \city{Martigny}
  \country{Switzerland}
}
\email{esenft@idiap.ch}

\author{Bilge Mutlu}
\orcid{0000-0002-9456-1495}
\affiliation{%
  \institution{University of Wisconsin--Madison}
  \city{Madison}
  \state{Wisconsin}
  \country{United States}
  \postcode{53706}
}
\email{bilge@cs.wisc.edu}


\begin{abstract}
  We introduce a taxonomy of important factors to consider when designing interactions with an assistive robot in a senior living facility. These factors are derived from our reflection on two field studies and are grouped into the following high-level categories: primary user (residents), care partners, robot, facility and external circumstances.
  We outline how multiple factors in these categories impact different aspects of personalization, such as adjusting interactions based on the unique needs of a resident or modifying alerts about the robot's status for different care partners. 
  This preliminary taxonomy serves as a framework for considering how to deploy personalized assistive robots in the complex caregiving ecosystem.
\end{abstract}

\ccsdesc[500]{Computer systems organization~Robotics}
\ccsdesc[300]{Social and professional topics~Socio-technical systems}
\ccsdesc[300]{Social and professional topics~Seniors}
\ccsdesc[100]{Human-centered computing~Field studies}

\keywords{Human-robot interaction, older adults, caregiving, assistive robots, taxonomy, personalization}



\maketitle

\section{Introduction}
\begin{figure}[!t]
    \centering
    \includegraphics[width=\columnwidth]{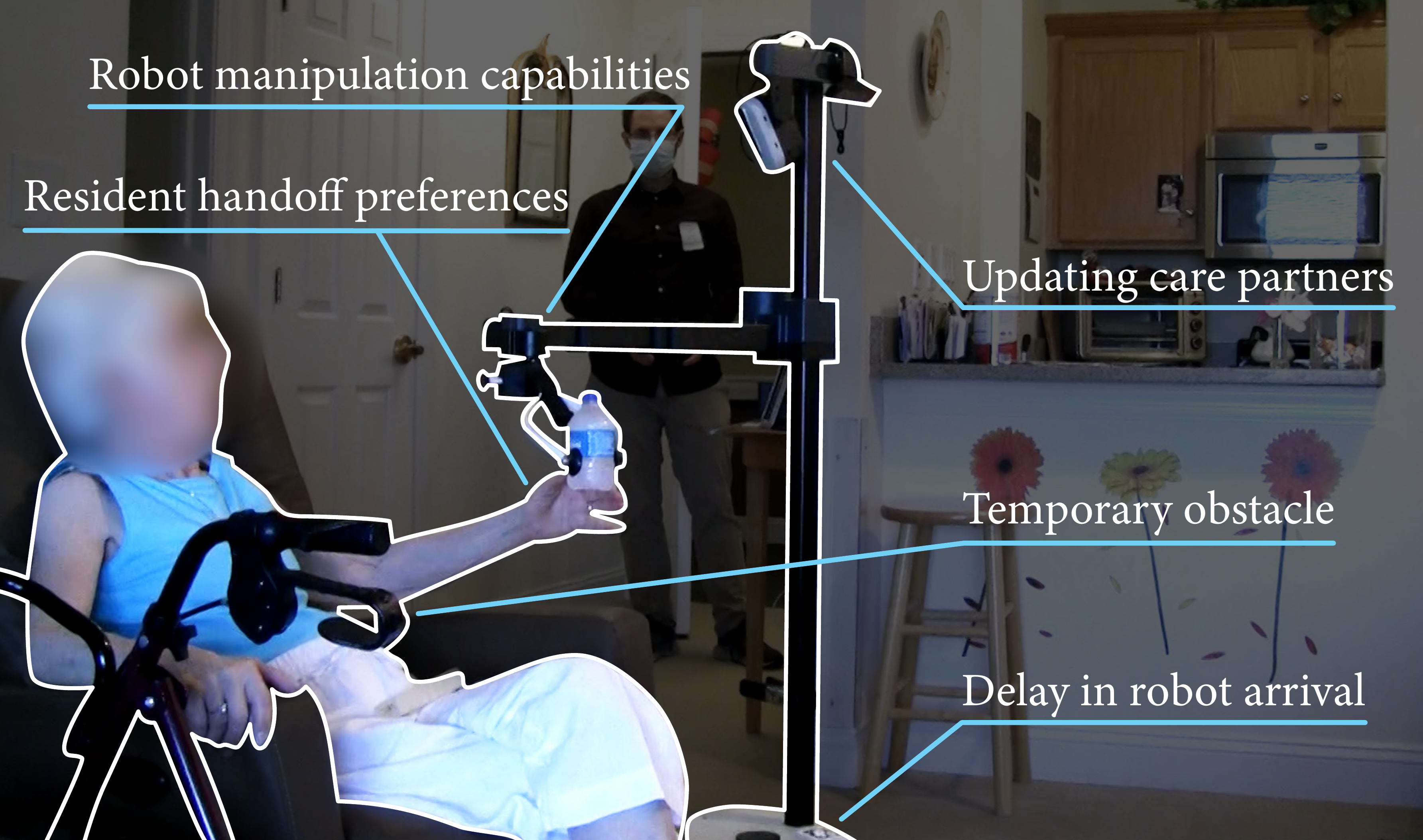}
    \caption{Designing robots to assist residents in senior living facilities necessitates considering a wide range of factors, which we introduce as a taxonomy to guide future researchers. This photograph from one of our field studies is annotated with some considerations for robot design that are based on our taxonomy.}
    \label{fig:teaser}
\end{figure}

Nations around the world face challenges related to an increasingly aging population \cite{lutz2008coming}. For example, the United States is estimated to have a shortage of 355\thinspace000  caregivers by 2040 \cite{famakinwa2021report}. Due to this shortage, some individuals may not be able to access resources for healthy aging. To address this challenge, recent research has focused on technology that can supplement care (\textit{e.g.}, ambient assisted living---AAL \cite{cicirelli2021ambient}). Robots, such as the one pictured in Figure \ref{fig:teaser}, have shown especially great promise in assisting with care-related activities \cite{abou2020systematic}.
%

Assistive robots are being developed to help with not only care tasks that are more social in nature, such as providing medication reminders \cite{prakash2013older}, comfort \cite{chen2022can}, and social stimulation \cite{luperto2019towards}, but also with physical care tasks such as refilling water \cite{odabasi2022refilling}, helping with ambulation \cite{mederic2004design}, bathing \cite{king2010towards}, monitoring and promoting safety \cite{gross2015robot}, and escorting residents to activities \cite{pollack2002pearl}. 

Even tasks perceived as more mechanical in nature could require social interaction capabilities and personalization due to interactions with \textit{ residents} of the care facility and the facility staff, such as \textit{ caregivers}. For example, \citet{odabasi2022refilling} reported that residents wanted to speak to the robot and caregivers requesting communication from the robot about what it was doing. In addition to these social skills, adapting to individual needs will be important for the overall success of these systems. \citet{pollack2002pearl}, for example, describes that robots should learn and adapt to the residents' preferred walking speeds so that they can move at an appropriate pace.

While human-robot interaction (HRI) with older adults has been widely studied, a majority of the work has focused on specific facets in isolation, such as the robot's appearance \cite{broadbent2009retirement}, acceptance of the robot \cite{alaiad2014determinants}, which tasks the robot should perform \cite{smarr2012older}, and technical ability to achieve the task \cite{odabasi2022refilling}.
Other HRI work not specific to older adults has uncovered varying preferences in how robots should approach people \cite{dautenhahn2006how} or even hand off objects \cite{choi2009hand}. These components together contribute a wealth of knowledge toward the successful design and development of assistive robots, yet bigger-picture considerations for how these robots can fit into the existing caregiving ecosystem are still underdeveloped. 

The existing caregiving ecosystem is complex, even before the introduction of robots. Caregivers have existing workflows they wish to maintain and residents have specific routines and preferences they want to be honored. Conflicting opinions on care can easily cause tension between residents, their families, and their caregivers. Care facilities have policies and customs that dictate behavior \cite{garrod2020advanced}, and even the physical layout can impact interpersonal behavior \cite{zimmerman2001assisted}. Facilities are regulated by government laws and agencies. All of these factors need to be considered when designing assistive robots, and they all need to be adjusted based on individual circumstances.
\citet{mois2020role} introduce a socio-ecological model that encompasses these factors at a high level by discussing the challenges of assistive robots across four nested levels: individual residents, care partners, community healthcare, and state/federal healthcare system. This descriptive model focuses on organizational needs to ensure successful adoption of assistive robots, such as ensuring clinical effectiveness, adapting to ever-changing user needs, and considering technology competency. We seek to build off of this framework by offering a finer-grained perspective into the day-to-day interaction considerations that impact ecological fit.

This paper contributes a preliminary taxonomy of factors to consider to integrate robots into the caregiving ecosystem. We first describe the studies that led to these factors, then present the taxonomy in detail, and close with a discussion of how the taxonomy can guide personalization efforts.



\section{Method}
We introduce this taxonomy as a preliminary set of factors based on the findings of and reflection on two previous field studies in a senior living facility. Through this past work, we have immersed ourselves and our robot in the caregiving ecosystem. Then we reflected as HRI researchers on what factors we should consider for future robot deployments.

The first study, detailed in our work \citet{stegner2022designing}, used ethnographic and co-design methods to better understand caregiver workflows and envision how assistive robots can fit into and support day-to-day caregiving tasks. We shadowed caregivers during their shifts to see their care practices in context. Then, after the shadowing was completed, several caregivers participated in individual co-design sessions where they discussed challenges of their work and brainstormed how a robot could help. The results included detailed workflows of caregivers during their shifts, as well as sketches created by the caregivers that illustrated what capabilities they would require from a robot to properly assist them with their work. Overall, the results call attention to the need for consideration for a triadic interaction between the residents, robots, and caregivers (not just residents and robots). Caregivers need to be able to convey their knowledge of the residents' needs and preferences to the robot, and the robot also needs to fit into the existing control hierarchy between caregivers and residents.

The second study, detailed in our work \citet{stegner2023situated}, considers designing assistive robots from the perspective of the residents. This study employs a new method called \textit{Situated Participatory Design}, which we introduce in our paper, to better understand how residents want to interact with a robot in their day-to-day lives. The three-phase study involved (1) co-design sessions with each resident to introduce the robot and design and enact a task that the robot could complete for that resident based on their individual needs, (2) simulated deployments, where the interaction between the robot and resident was tested under realistic conditions and iteratively updated according to resident feedback, and (3) follow-up interviews with caregivers to reflect on the resulting designs. For this study, we worked with the Stretch RE1, which is a mobile manipulator robot suitable to navigate a home environment and grab light objects \cite{kemp2022design}. The results included concrete designs where the robot would assist the resident with a variety of tasks, such as delivering a newspaper/mail/library book/cup of ice/water bottle, picking up a towel from the floor, or moving a cup of water across the room. Generally, residents were happy to receive assistance from the robot, and the caregivers were enthusiastic that these tasks would be suitable for the robot to complete. We found that residents had a variety of preferences and needs for how the robot should behave, and these desires changed over time based on both further exposure to the robot and also changes in their day-to-day mood or condition.

Each of these studies provided a different but connected perspective on how the robot should behave. In addition to considering the personalized interaction between the robot and the resident, we also need to consider other ecological factors, such as the how the needs of the caregivers affect the robot's behaviors or how the facility layout and robot design impact the robot's ability to complete specific tasks.
Given the complex nature of this space, we reflect on our experience to generate an initial taxonomy of factors that we perceived to impact robot interaction design.

\section{Taxonomy}
This preliminary taxonomy provides insight into various ecological factors that impact the personalization of assistive robots in a senior living facility. The organization is as follows:
\begin{enumerate}
    \item A breakdown of different components of an interaction
    \item A taxonomy describing factors that can impact the interaction, divided into five main categories.
\end{enumerate}
This taxonomy is not intended to be exhaustive, but it is instead serves as a starting point to guide future researchers when thinking about designing robotic interactions.

\subsection{Interaction components}
We first break down an interaction into components that are primarily based on the temporal flow/progression through a task. These components are important because different factors of our taxonomy influence different interaction components:
\begin{enumerate}
    \item \textit{Task selection} --- Deciding which task the robot will perform. Task selection is about \textit{what} the robot will do, not when it would do so or how to start. This component could happen immediately prior to the task or further in advance. The task could also be selected by different stakeholders, such as a caregiver sending the robot to check on a resident or the resident calling the robot for help with something.
    
    \item \textit{Prior to the task} --- Before the robot does the task, it may need to do some preparation such as asking permission from a caregiver or logging the task it plans to complete.
    
    \item \textit{Task initiation} --- When the task should begin, something needs to initiate it. For example, the resident could call the robot, the caregiver could send the robot, or the robot could start at a pre-arranged time.
    
    \item \textit{Entering the space} --- If the robot needs to transition between spaces, such as from a public common space to the resident's private room, how it will do so (\textit{e.g.}, knock and wait for an answer before entering).
    
    \item \textit{During the task} --- What the robot should do while it is completing the task. For example, the robot could deliver the mail by placing it in a pre-defined location or be prepared to listen for the resident to specify where to place the mail once it arrives to the room.
    
    \item \textit{After the task} --- Once the robot has completed the task and finished the direct interaction with the resident, it may need to do something else, such as update a log or send a status report to family.

    \item \textit{Social elements} --- The social behavior the robot should include during the interaction, such as light chit chat or more involved conversation.
    
    \item \textit{Evolution over time} --- Interactions could change from day to day, either due to temporary situations (\textit{e.g.}, a visitor or a special event) or shifts in preference or needs (\textit{e.g.}, a change in daily routine or a new dietary restriction).
\end{enumerate}

Although these interaction components are contextualized in caregiving for older adults, most human-robot interactions follow similar patterns, from more controlled lab studies to other in-the-wild scenarios such as in hospitality or education. 

\subsection{Factors}
\begin{figure}
    \centering
    \includegraphics[width=\columnwidth]{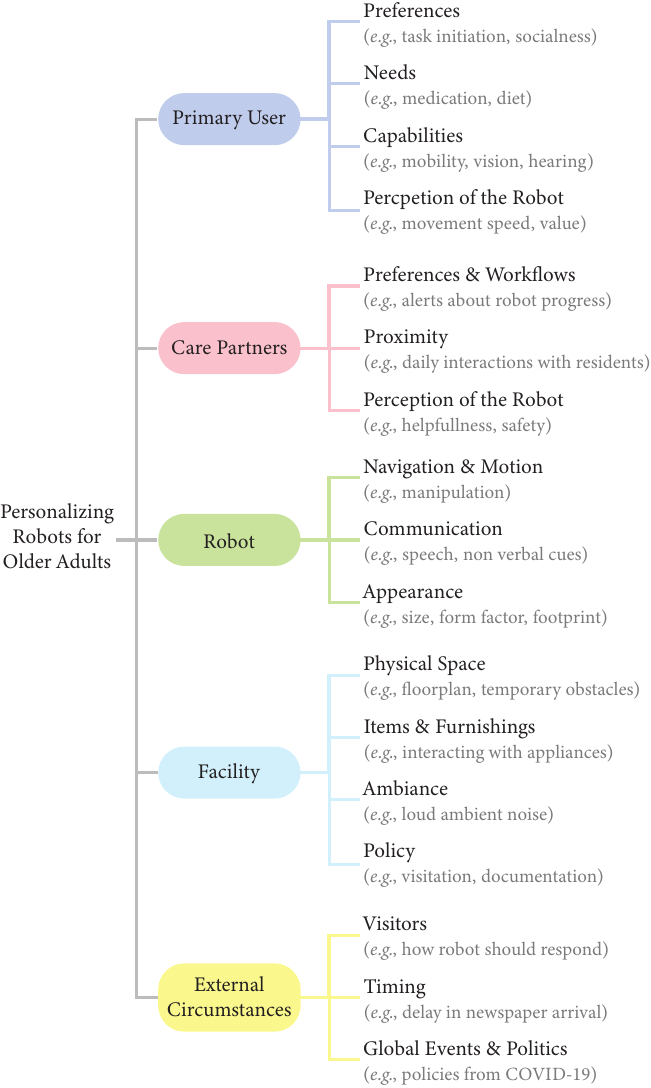}
    \caption{Factors to consider when personalizing a robot within the caregiving ecosystem, grouped into five categories.}
    \label{fig:taxonomy}
\end{figure}

We now present our taxonomy of factors that impact the interaction outlined above. Each factor of the taxonomy, which is outlined in Figure \ref{fig:taxonomy}, represents a major consideration for designing interactions with assistive robots for older adults. The factors are organized into higher-level categories based on their source, including various actors (\textit{e.g.}, residents, care partners, the robot) or the environment (\textit{e.g.}, the facility, external circumstances).
Some factors include numerous sub-components that provide further granularity toward understanding how they can impact the interaction.

\subsubsection{\textbf{Primary User (Resident)}}
The resident may be considered the ``primary user'' of the robot, as they are the care recipient for the robot's assistance. Residents can shape interactions not only based on their personal preferences and needs, but also through their physical/mental capabilities and perceptions of the robot.

\textbf{Preferences --- } 
Personal preference greatly impacts how the robot should interact with residents. We found that residents had preferences on the task initiation (\textit{e.g.}, resident-initiated, pre-set time, flexible time based on external events), entering the space (\textit{e.g.}, knock and enter; knock, make an announcement, and enter; knock and wait for a response before entering), socialness of the robot (\textit{e.g.}, limited verbal interaction, light chit chat, or even complex conversation), and during the task (\textit{e.g.}, check for  other tasks to do, wait for instructions of what to do, say goodbye before leaving).

\textbf{Needs --- }
Residents may also have specific needs that must be met due to circumstances such as:
\begin{itemize}
    \item \textit{Medication} --- Residents may take medication to help manage various health conditions. We witnessed one case where a resident's dexterity and mobility fluctuated significantly whether or not they had recently taken their medication. This variance can mean unpredictable changes in a resident's needs and abilities during an interaction.
    \item \textit{Diet} --- Residents may also need to adhere to special diets, such as limiting sugar intake in the case of a resident with diabetes or requiring thickened drinks in the case of a resident who has difficulty swallowing. Failure to meet these dietary needs can cause significant harm to residents.
    \item \textit{Other Assistance} --- Residents often require assistance with daily care activities such as using the toilet, transferring between a bed and chair, bathing, or getting dressed. They may also be in need of emotional or social support. These varying care needs can be reflective of the resident's capabilities, which we discuss next.
\end{itemize}

\textbf{Capabilities --- } 
The physical capabilities of a resident dictate their ability to interact in certain ways:
\begin{itemize}
    \item \textit{Mobility} --- Mobility can impact a resident's ability to move around their living space, which increases the need for a mobile robot that could enter and exit on its own or move through the living space to find and interact with the resident where they are. For example, one of our participants noted that if they had to get up to open the door for the robot, then it would not actually offer them any benefit.
    \item \textit{Dexterity} --- A resident with limited dexterity could struggle to physically interact with the robot, making a touch input infeasible (\textit{e.g.}, tablet touch screen). For example, one participant previously had a stroke and, when combined with arthritis symptoms, was not able to use either hand for fine motor skills.
    \item \textit{Vision} --- A resident with a vision impairment may require the robot to move very close so that it can be seen. For example, one of our participants had a vision impairment, so they always wanted the robot to approach extremely closely and would often reach out to touch it to gain a better understanding of it.
    \item \textit{Hearing} --- If a resident cannot hear well, they may need the robot to be extremely loud or communicate in a way that does not rely on auditory input. For example, one of our participants normally wore a hearing aid, but they would occasionally not wear it or it would be out of batteries, so they would not be able to hear the robot's speech. 
    \item \textit{Speech} --- Some residents cannot speak clearly, which can make voice detection nearly impossible with current technology. Residents who cannot be understood by the robot would need an alternative means to provide input, such as a touch screen or gestures.
\end{itemize}

\textbf{Perception of the Robot --- }
Residents had different perceptions of the robot either from their general notion of robots or from their direct experience in our study. Their perceptions influenced the kind of interaction they desired with the robot. We observed the following perceptions:
\begin{itemize}
    \item \textit{Movement Speed} --- Our robot moved quite slowly, causing some residents to require verbal updates of its progress while it moved through hard-to-see areas of their home.
    \item \textit{Robot Size} --- Some residents felt the robot was a good size, because it was small and maneuverable. One resident, however, used a wheelchair and felt that the robot was too tall for them to interact with.
    \item \textit{Concern for Personal Belongings} --- Two residents felt concern that the robot could mishandle their personal belongings. One was concerned with the security of the robot handling the mail, which ruled that out as a task they were comfortable with the robot doing. Another was concerned the robot may knock things over while completing the task, and therefore wanted to help the robot instead of the robot doing everything on its own.
    \item \textit{Value} --- Some residents valued having the robot complete some tasks because they felt they were otherwise a burden to human caregivers, while others preferred the caregivers due to a lack of trust in the robot's abilities.
\end{itemize}

\subsubsection{\textbf{Care partners}}
Care partners include other stakeholders who help contribute or monitor the resident's care, such as caregivers, family, friends, or other care facility staff.

\textbf{Preferences and Workflows --- } Care partners, particularly caregivers and heavily involved family members, may have preferences based around their workflows. For example, some caregivers expressed that the robot should not alert them for every small update because they have quite a bit of work to do and do not want unnecessary notifications. However, a relative who is concerned about care may want more frequent updates from the robot.

\textbf{Proximity --- } Care partners who are in close proximity to the resident (\textit{e.g.}, caregivers who see them daily) may desire a different level of supervision and updates about the robot's activities compared to a family member who is involved in the care plan but geographically far away or unable to visit.

\textbf{Perception of the Robot --- } The tasks that a caregiver or family member could trust a robot to do will depend on their perception of it. For example, many caregivers we spoke with were optimistic and excited about the prospect of robotic assistance. However, several expressed concerns about whether the robot would be able to safely handle complicated scenarios such as ensuring dietary requirements are met for any food the robot may provide to a resident.

\subsubsection{\textbf{Robot}}
The the autonomous or semi-autonomous agent that provides assistance to residents. Robots have various capabilities that will impact what they can do and how they can interact. Some characteristics are fixed based on the robot's physical design and hardware, but others could be modified through software changes.

\textbf{Navigation \& Motion --- }
The robot's ability to move and interact with its environment depends on its navigation and grasping capabilities. The sensors and actuators on the robot, such as an RGB-D camera and LIDAR sensor for navigation or actuated arm for manipulation, can dictate which tasks it is able to do. The simple Cartesian arm on our robot was limited from more complex manipulations but still able to complete simple tasks that both residents and caregivers viewed as useful.

\textbf{Communication --- }
A robot can have various features to facilitate communication, including a microphone to perceive speech and a speaker to project a response. Our robot was not equipped with a powerful enough speaker to process speech input or play a response loudly enough that participants with hearing impairments could hear. The text-to-speech engine had to be updated to a deeper, slower voice to accommodate some of the residents' capabilities. To supplement verbal communication, the robot may signal non verbally, such as using LED lights or making facial expressions. We found some residents wanted the robot to use arm gestures to signal if it was listening, or \textit{thinking}, about what to say.

\textbf{Appearance --- }
What the robot looks like, including its size, form factor, and footprint. The appearance can impact comfort around and acceptance of the robot. For example, while many of our participants appreciated the small footprint of our robot, one resident felt too uncomfortable in the presence of the robot and withdrew from the study shortly after seeing it.

\subsubsection{\textbf{Facility}}
The facility refers to the physical location where the resident lives. The facility's floorplan, furnishings, and atmosphere will dictate what the robot can do and how it will do so.

\textbf{Physical Space --- } 
The physical layout of a care facility can impact where the robot is able to go and what abilities it needs to function successfully. For example, many robots cannot climb stairs, and a robot without arms would need external help to press elevator buttons. In this case, a robot may be confined to one floor. In addition to permanent blocks, the robot may encounter temporary obstacles such as a medicine cart or resident blocking its path. These blockages could impact the timing of when a robot can complete certain tasks or how quickly it is able to move from space to space.

\textbf{Items \& Furnishings --- } 
The items and furnishings within a space affect what the robot can do. For example, we experienced that our robot struggled to open the refrigerator in the residents' rooms, making tasks involving that activity likely to fail. We also had the opportunity to have the robot interact with residents' personal items such as a newspaper or cup, to see which manipulations were feasible and which would require hardware modifications.

\textbf{Ambiance --- }
The atmosphere of the space can impact the robot's ability to interact. We experienced several instances of loud ambient noise, such as live music from a sing along in a nearby public common space or from loud televisions playing in the rooms of residents who are hard of hearing.

\textbf{Policy --- }
The facility may have specific policies such as rules regarding visitation, access to certain spaces, and care documentation. Some of these policies may be implemented to comply with laws and regulations, whereas others may be a result of workplace culture. Since these policies can vary across facilities or even within different areas within the same facility, robots need to be able to adapt to ensure proper compliance.

\subsubsection{\textbf{External Circumstances}}
The caregiving ecosystem is not self contained and is instead influenced by external factors that are not controllable by the researcher/designer.

\textbf{Visitors --- }
Residents might have visitors when the robot arrives, which can change the interaction. The desired robot behavior (\textit{e.g.}, engage in a multi-party interaction, wait, or leave) will have to be personalized based on that resident's preference, which could be influenced by the nature of the visitors. For example, once our robot entered while a resident had a friend visiting, and the robot became a new topic of shared interest. However, in the case of a nurse or social worker visiting, the resident may prefer the robot to wait or return at a later time.

\textbf{Timing --- }
The timing of external events, such as mail delivery, is not consistent. Weather, traffic, or other unforeseen circumstances could easily create delays in outside operations. One of our participants changed the tasks they wanted to do from a newspaper delivery to something else because they wanted to know when to expect the robot and the newspaper would often arrive late.

\textbf{Global Events and Politics --- }
Global events can impact care facilities. We experienced significant disruption due to the COVID-19 pandemic, including temporary halts to our work. Requirements for robot sanitation can impact the industrial design, which can constrain its appearance and functionality. Although these events cannot be anticipated, researchers and designers need to be aware that they will need to have contingency plans in such situations.

\section{Discussion}
The above taxonomy presents a preliminary set of factors that we found to impact the day-to-day interactions of an assistive robot. 
We propose this taxonomy as we believe other researchers could use it as a starting point when designing robots for older adults. Some of these factors reflect what the robot can use to personalize its behavior (\textit{e.g.,} voice or motions) while others reflect the basis for adapting the robot's behavior (\textit{e.g.,} preferences and needs). We plan to carefully consider these factors in our future participatory design studies and deployments, for example by assessing early in the project what the participants' capabilities are and using that information to tailor our questionnaires and interviews accordingly. Furthermore, these factors will help us in designing interfaces that we can provide to users to allow them to personalize the robot's behaviors directly. Although these factors emerged from a caregiving scenario, we anticipate very similar ones would exists in other environments such as schools or museums.


\subsection{Dynamic Personalization}
Altough most of the factors appear static, such as the parts of a facility accessible to the robot or regulations that impact day-to-day operations, in practice all factors are subject to change. This change can be sudden, temporary, or gradual. For example, while the floor plan of the facility may be relatively fixed, maintenance could temporarily or permanently alter which spaces the robot is able to access. Similarly, the high rate of caregiver turnover \cite{castle2020nurse} can lead to sudden changes in care partner needs and preferences. When designing personalized robots, we need to keep in mind that personalization is not a one-step effort, but an on going process.

Residents themselves are not fixed, and their preferences and capabilities can change from day to day. In our work, residents interacted with the robot multiple times across multiple different days. While some had relatively static needs and preferences, others were much more dynamic. In several scenarios, the residents were adamant about not needing to speak to the robot, but in the end, when actually interacting with the robot in a more natural way, they wanted the robot to stay to chat that day because they desired to get to know it more. 
To accommodate such changes, personalization should be viewed as a dynamic approach to interactions, rather than a fixed configuration to set and forget.

\subsection{Perfect Personalization --- A Fool's Errand}
Another point that emerged from our design studies is that we cannot design a perfect system that can fit every user's expectations. 
Sometimes it may be possible to personalize the interaction according to all parties' needs and preferences, but in other cases these factors can compete.
For example, there is no conflict when personalizing the timing and frequency of the robot's activity updates for different care partners. However, residents can have preferences on how the robot should enter their private space that could conflict with the facility's policies about what protocol staff must follow. While human caregivers may deviate from the protocol due to a personal connection with the residents, the robot may need to more strictly adhere to these policies and thus breach some of the personalization efforts. Similar conflicts exist within the caregiving ecosystem already, but introducing robots and other technologies means that these implicit arrangements need to be made explicit. 

Even if the resident and care partners agree on the interaction, the robot may still face challenges with the physical environment or its capabilities. For example, if the robot is unable to reach a high shelf, no amount of software personalization can solve this issue. Instead, either the environment or robot hardware would need updated, which is possible but might be costly and invasive.
Personalization can be limited by such factors, which are important to consider as they are readily present in real-world scenarios.

Due to these limitations, it is important to consider which aspects are more critical (\textit{e.g.}, appropriately communicating with residents based on their capabilities) and which aspects are more flexible (\textit{e.g.}, how to treat visitors). There also remains a level of personalization of the users themselves --- as long as the robot's capability are clearly described and the users' expectations are set appropriately, we can rely on people's ability to adapt to the robot too.

\subsection{Limitations \& Future Work}
Our taxonomy is based on data collected at only one facility. While we expect that it will generalize to other care facilities, future work is needed to provide additional support. 
In future work, we might strengthen this taxonomy by systematically reviewing literature on HRI for older adults in senior living communities to evaluate how other findings support, challenge, or enrich our taxonomy. Incorporating such literature review would extend consideration beyond our observations to a broader discussion on which aspects are currently addressed and which are not yet undertaken. Similarly, we could also conflict our taxonomy to other use cases (such as education or hospitality) to make it more generalizable.

\begin{acks}
We thank the residents and caregivers for their participation. 
This material is based upon work supported by a University of Wisconsin--Madison Vilas Associates Award, National Science Foundation (NSF) award IIS-1925043, and an NSF Graduate Research Fellowship under Grant No. DGE-1747503. 
Any opinions, findings, and conclusions or recommendations expressed in this material are those of the authors and do not necessarily reflect the views of the NSF.
\end{acks}

\bibliographystyle{ACM-Reference-Format}
\bibliography{biblio}

\end{document}